\documentclass[10pt,journal,letterpaper,compsoc]{IEEEtran}
\usepackage[breaklinks=true,colorlinks]{hyperref}

\usepackage[nocompress]{cite}
\usepackage{algorithm}

\usepackage{subfigure}
\usepackage{url}
\usepackage{multicol,multirow}
\usepackage{amssymb}
\usepackage{amsmath}
\usepackage{ragged2e}
\usepackage{overpic}
\usepackage{rotating}
\usepackage{color}
\usepackage{xcolor}
\usepackage{etoolbox}
\makeatletter \patchcmd{\@makecaption}{\\}{.\ \justifying }{}{}\makeatother
\usepackage{graphicx}
\DeclareGraphicsExtensions{.pdf,.jpg,.png}
\usepackage{booktabs}
\usepackage{amsfonts}
\usepackage{graphicx}
\usepackage{subfigure}
\usepackage{multirow}
\usepackage{amssymb}
\usepackage{amsmath}
\usepackage{diagbox}
\usepackage{bbding}
\usepackage{makecell}
\usepackage{xcolor}
\usepackage{float}  
\usepackage{listings}

\lstdefinestyle{mystyle}{
  backgroundcolor=\color{white},   
  commentstyle=\color{codegray},
  keywordstyle=\ttfamily\scriptsize\color{codeblack},
  numberstyle=\tiny\color{codegray},
  basicstyle=\ttfamily\scriptsize\color{codeblack},
  breakatwhitespace=false,         
  breaklines=true,                 
  captionpos=b,                    
  keepspaces=false,                
  showspaces=false,                
  showstringspaces=false,
  showtabs=false,                  
  tabsize=2
}

\renewcommand{\tablename}{Table }

\renewcommand{\eqref}[1]{Eq.~\ref{#1}}

\newcommand{\myPara}[1]{\vspace{10pt}\noindent\textbf{#1.}\quad}

\usepackage{xcolor}         
\definecolor{codegreen}{rgb}{0,0.6,0}
\definecolor{codegray}{rgb}{0.501961,0.501961,0.501961}
\definecolor{codepurple}{rgb}{0.58,0,0.82}
\definecolor{codeblack}{rgb}{0,0,0}

\RequirePackage{silence}
\graphicspath{{./figures/}}

\begin{document}

\title{PVP: Pre-trained Visual Parameter-\\Efficient Tuning}
\author{Zhao Song, Ke Yang*, Naiyang Guan*, Junjie Zhu, Peng Qiao, and Qingyong Hu
\IEEEcompsocitemizethanks{
\vspace{-5pt}
\IEEEcompsocthanksitem Z. Song, K. Yang, N. Guan, and J. Zhu are with the Defense Innovation Institute, Beijing, China.
\IEEEcompsocthanksitem P. Qiao is with  the National University of Defense Technology, Changsha, China.
\IEEEcompsocthanksitem * denotes the corresponding author.
}
}

\IEEEcompsoctitleabstractindextext{%
\begin{abstract}
\justifying
Large-scale pre-trained transformers have demonstrated remarkable success in various computer vision tasks. However, it is still highly challenging to \textit{fully fine-tune} these models for downstream tasks due to their high computational and storage costs. Recently, Parameter-Efficient Tuning (PETuning) techniques, \textit{e.g.}, Visual Prompt Tuning (VPT) \cite{vpt} and Low-Rank Adaptation (LoRA) \cite{Lora}, have significantly reduced the computation and storage cost by inserting lightweight prompt modules into the pre-trained models and tuning these prompt modules with a small number of trainable parameters, while keeping the transformer backbone frozen. Although only a few parameters need to be adjusted, most PETuning methods still require a significant amount of downstream task training data to achieve good results. The performance is inadequate on low-data regimes, especially when there are only one or two examples per class. To this end, we first empirically identify the poor performance is mainly due to the inappropriate way of initializing prompt modules, which has also been verified in the pre-trained language models. Next, we propose a Pre-trained Visual Parameter-efficient (PVP) Tuning framework, which pre-trains the parameter-efficient tuning modules first and then leverages the pre-trained modules along with the pre-trained transformer backbone to perform parameter-efficient tuning on downstream tasks. Experiment results on five Fine-Grained Visual Classification (FGVC) and VTAB-1k datasets demonstrate that our proposed method significantly outperforms state-of-the-art PETuning methods. As highlighted below, we show that our PVP framework achieves 16.08$\%$, 11.52$\%$, 6.36$\%$, 2.94$\%$, and 1.95$\%$ average accuracy improvement under 1, 2, 4, 8, and 16 shot setting on FGVC, respectively, compared with the previous PETuning techniques, \textit{e.g.}, VPT, in the task of few-shot image classification. PVP also achieves state-of-the-art results in the VTAB-1k benchmark, surpassing the average accuracy of very recent PETuning methods by 2.33$\%$.
\end{abstract}

\begin{IEEEkeywords}
	Parameter-Efficient Tuning, Prompt Tuning, Vision Transformer, Few-shot Learning, Transfer Learning.
\end{IEEEkeywords}
}

\maketitle
\IEEEdisplaynotcompsoctitleabstractindextext
\IEEEpeerreviewmaketitle

\section{Introduction} \label{intro}

\IEEEPARstart{I}{n} the past few years, vision transformer models including ViT~\cite{vit} and Swin~\cite{swin}, have achieved encouraging results on a number of mainstream vision tasks. However, training such large transformer models is usually accompanied by massive training data and expensive computational costs, making it highly challenging for individuals to train such models from scratch. Fortunately, the industry technology giants including Microsoft and Facebook, have released models with carefully pre-trained parameters on large-scale pre-training data~\cite{russakovsky2015imagenet}, enabling individuals to use large transformer models by either fine-tuning all the model parameters or just a small proportion of model parameters~\cite{2014How,2016Unsupervised,sidetune,BitFit,adapterfusion,adapterhub} while keeping the majority frozen.

\begin{figure}
    \centering
    \includegraphics[width=1\linewidth]{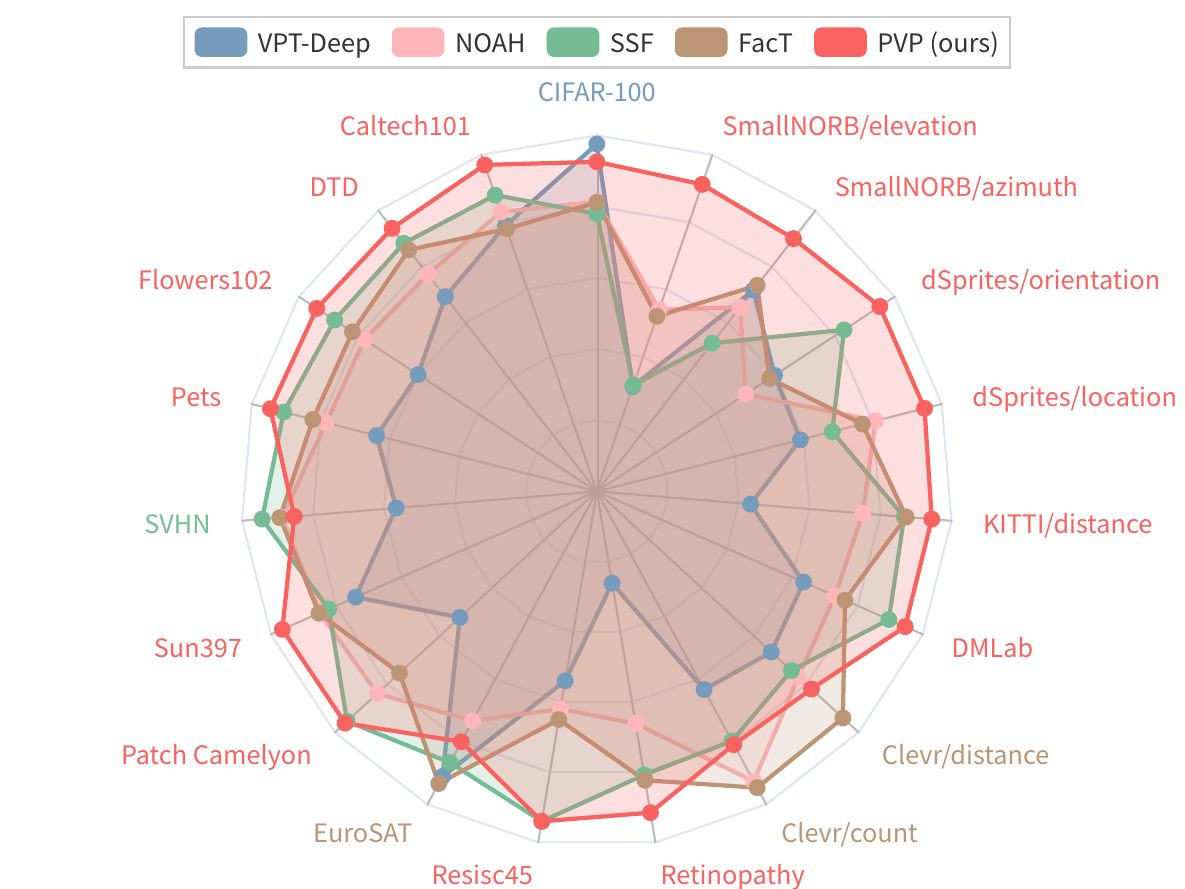}

    \caption{Our proposed PVP demonstrates  strong performance over recent state-of-the-art methods on the VTAB-1k benchmark. The dataset names are color-coded to indicate the best-performing method for each dataset clearly.
    }
    \label{fig:radar_vtab}
\end{figure}

Recently, a handful of pioneering works termed Parameter-Efficient Tuning (PETuning) methods~\cite{vpt,adapter,NOAH,convbypass,SSF,FacT}, attempted to tune several newly inserted modules instead of part of the transformer backbone. For example, Visual Prompt Tuning (VPT)~\cite{vpt} is a PETuning method that adds task-specific learnable parameters, namely prompt tokens, to the input space and only fine-tunes the prompt tokens on downstream tasks. Notably, prompt tokens only account for less than 2$\%$ of total parameters. Intuitively, such a small amount of parameter adjusting is naturally suitable for the scheme of few-shot learning, where only a few data samples are provided for training. However, we empirically find that poor performance is achieved by VPT when limit tuning data (as shown in Sec. \ref{section:sec3.2}). In particular, the accuracy on the CUB-200-2011 dataset drops to 30.05$\%$ using 1$\%$ tuning data, compared to 88.50$\%$ accuracy using all tuning data. Motivated by this, we aim to explore the fundamental problems of why PETuning methods do not perform well on few-shot classification tasks.

We attribute this phenomenon to the inadequate initialization of prompt modules, since current PETuning methods, \textit{e.g.,} VPT~\cite{vpt}, Adapter~\cite{adapter}, and LoRA~\cite{Lora}, usually use zero- or random-initialized modules for PETuning, meaning the newly added modules need to learn from scratch on downstream tasks. Moreover, most PETuning methods require the insertion of trainable prompt modules at earlier layers, particularly at the beginning of the network, resulting in the weights of all later layers being scrapped. These two problems lead to the prompt module requiring a significant amount of data for training, which can prove challenging for downstream tasks. However, pre-training datasets, such as ImageNet~\cite{russakovsky2015imagenet}, offer ample data to meet the required training needs.

To this end, we propose a Pre-trained Visual Parameter-efficient (PVP) Tuning  framework. We first pre-train the newly added modules of PETuning on a large dataset and subsequently  leverage these pre-trained modules to perform PETuning on downstream few-shot learning tasks. The rationale behind our approach is that the pre-trained parameters offer an excellent foundation for PETuning, requiring only a few gradient updates to fine-tune the modules. The tuned modules can then be applied to tasks such as few-shot image classification. \textit{ Importantly, we note that the newly added tuning modules and the vision transformer backbone are pre-trained on the same dataset, hence no additional pre-training data is involved.}


In addition to its effectiveness in few-shot scenarios, our proposed Pre-trained Visual Parameter-efficient (PVP) Tuning approach is also applicable when sufficient tuning data is available. Our experimental results indicate that the module pre-training stage significantly improves the adaptability of the transformer backbone to downstream tasks, outperforming current PETuning methods. Thus, our approach represents a promising Parameter-Efficient method for large-scale pre-trained transformer models.

The proposed PVP framework can be readily applied to different PETuning methods, provided that these methods integrate tunable modules into the vision transformer backbone and tune the newly added modules while keeping the transformer backbone frozen during downstream task tuning. Specifically, our framework can be applied to methods such as VPT\cite{vpt}, Adapter\cite{adapter}, and LoRA\cite{Lora}. Our experimental results demonstrate that these methods experience significant performance improvements when augmented with our PVP.
 Our contributions can be summarized as follows.
\begin{itemize}
\setlength{\parsep}{0pt} 
\setlength{\topsep}{0pt} 
\setlength{\itemsep}{0pt}
\setlength{\parsep}{0pt}
\setlength{\parskip}{0pt}
\item To the best of our knowledge, this is the first study to clarify the limitations of Parameter Efficient Tuning (PETuning) techniques on few-shot tasks and tackle this issue by pre-training the  newly added PETuning modules.
\item We propose a simple yet efficient Pre-trained Visual Parameter-efficient (PVP) Tuning framework, which achieves significant performance gains on downstream few-shot tasks, particularly  in extremely low-data regimes with only 1 or 2 training samples per class. Our approach can be easily applied to various PETuning methods and achieves a great performance improvement. 
\item In addition to the few-shot scenario, our PVP approach also achieves state-of-the-art results on the Visual Task Adaptation Benchmark (VTAB-1k), outperforming recent PETuning methods by a large margin.
\end{itemize}

\section{Related Works}

\begin{figure}[t]
    \centering
    \includegraphics[width=1.0\linewidth]{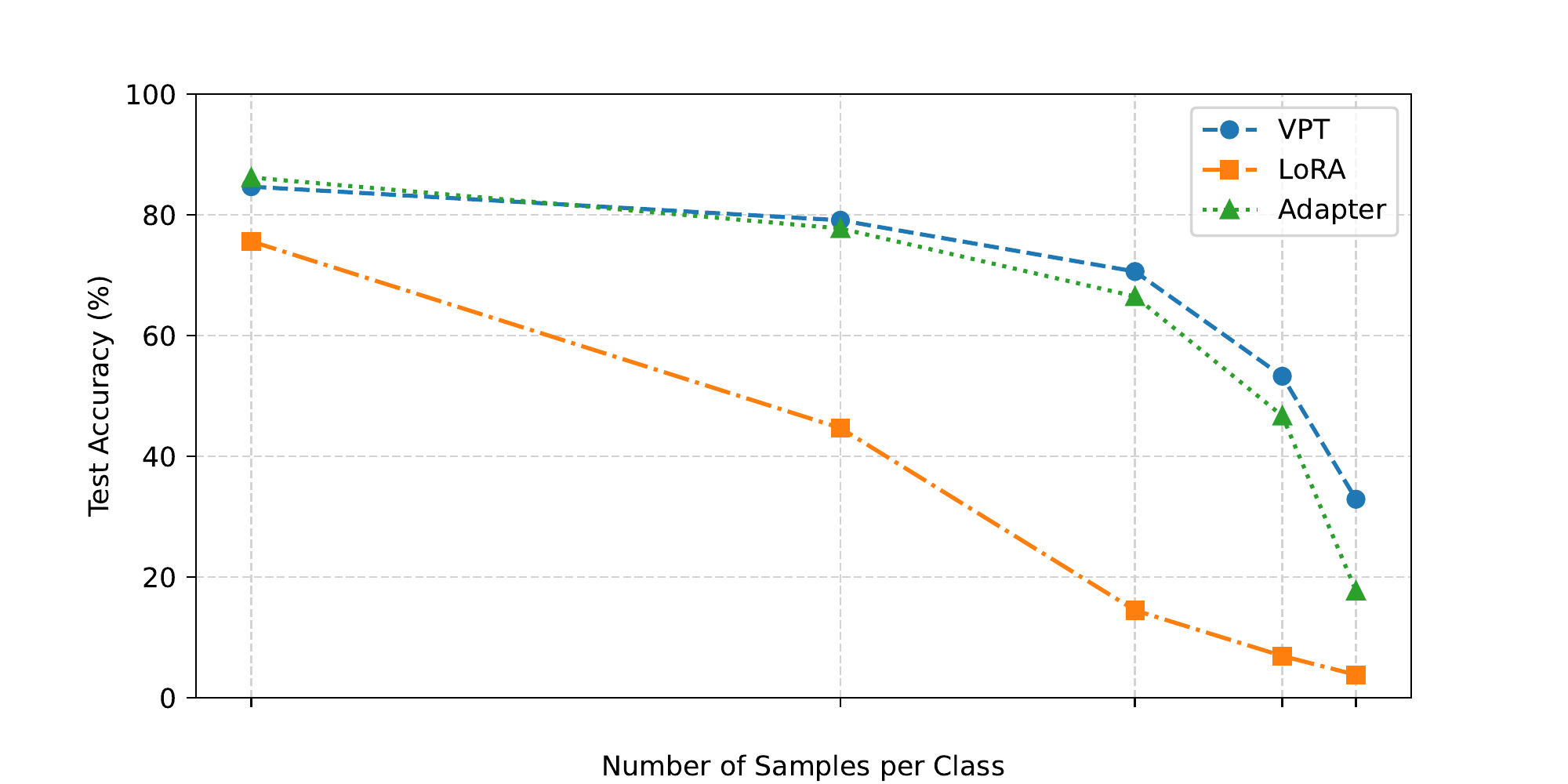}
    \caption{Performance degradation of existing visual PETuing techniques under few-shot classification setting on the CUB-200-2011 datasets.}
    \label{fig:prior_experiment}
\end{figure}

\subsection{Transformers In Vision}
Transformer is a type of deep neural network mainly based on self-attention mechanisms, which has been widely investigated due to its superior performance. Vaswani \emph{et al.}~\cite{transformer} proposed transformer architecture with a self-attention mechanism to capture the contextual relationship between inputs, and achieved great success in the Natural Language Processing (NLP) field~\cite{BERT,2020Language,2019Exploring,clip}. The remarkable success of large-scale transformer models in NLP has sparked a growing interest in adopting these models in Computer Vision (CV). Dosovitskiy \emph{et al.}~\cite{vit} introduced transformer architecture into the field of computer vision and proposed Vision Transformer (ViT). This is achieved by dividing an image into patches and then embedding these patches as tokens for the transformer encoder. Liu \emph{et al.}~\cite{swin} proposed swin transformer and calculated self-attention in the hierarchical local window while allowing cross-window interaction, which provided a multi-scale receptive field for the transformer. Subsequently, a variety of visual transformers~\cite{DeiT, CeiT, Localvit} are proposed to leverage knowledge distillation, convolutional embedding, and depth-wise convolution to improve the performance. Though vision transformer-based methods have achieved state-of-the-art performance in various vision benchmarks, fine-tuning pre-trained transformer models on downstream tasks is still data-dependent and computationally expensive, which limits the wider application of vision transformer models. Given that large-scale pre-trained models are publicly available, how to adapt the pre-trained transformers to downstream tasks~\cite{DBLP:conf/coling/HagstromJ22, DBLP:journals/corr/abs-2210-06989} in a parameter and memory efficient way remains a crucial open problem.

\begin{figure*}[t]
    \label{fig:ppetuning}
    \centering   
    \includegraphics[width=0.95\linewidth]{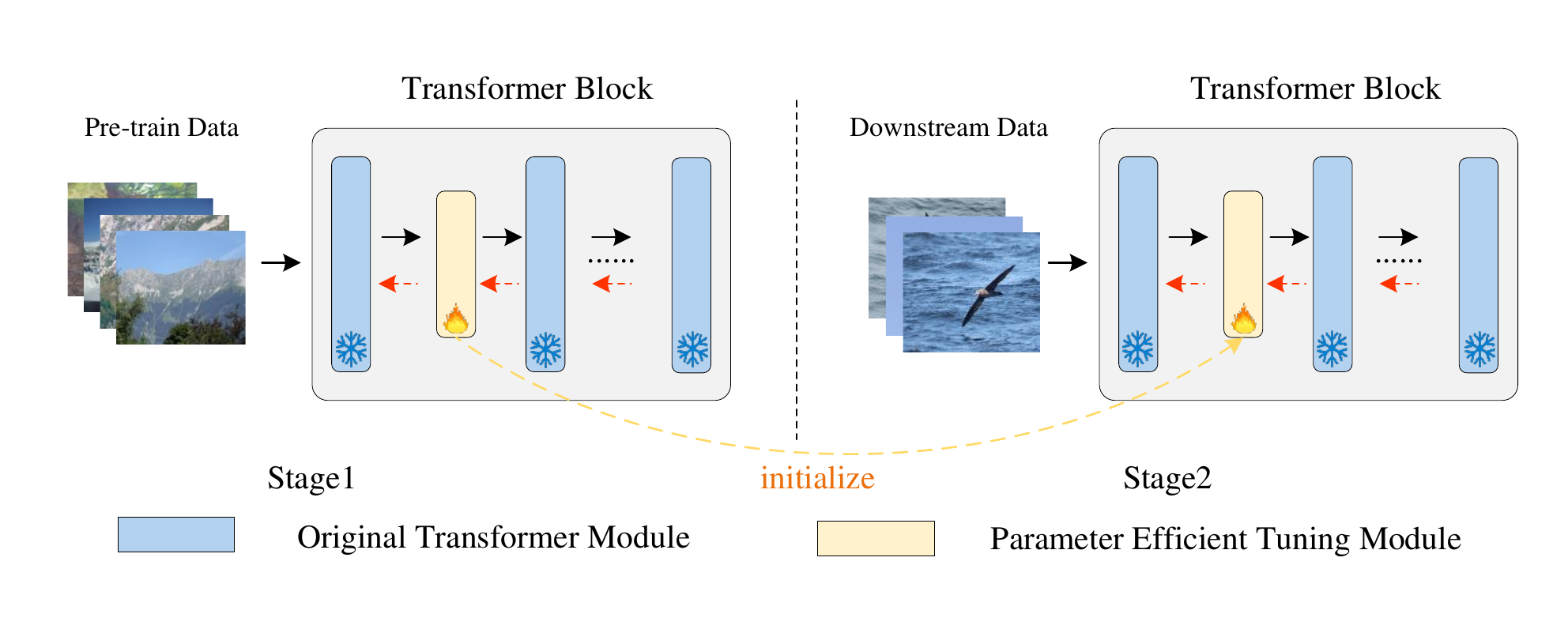}
    \caption{Overview of Pre-trained Parameter Efficient Tuning. There are two stages for our pre-trained parameter efficient tuning method. (1) Parameter Efficient Tuning module pre-train stage and (2) Downstream Parameter Efficient Tuning stage. Original transformer modules are frozen and parameter-efficient tuning modules are tunable in both stages. The learned parameter efficient tuning modules in stage 1 are used to initialize these in stage 2. The black and red rows represent forward and backward respectively.}
\end{figure*}

\subsection{Parameter Efficient Tuning}
The past few years have witnessed the huge success of parameter-efficient transfer learning in NLP~\cite{PES_LLMFT, DBLP:conf/coling/ZhouMZCGZHXW22, DBLP:journals/corr/abs-2210-16771, LadderSideTune, UniPELT, DeltaTuning}. Recently, parameter-efficient tuning methods on the pre-trained vision transformer models have been widely explored. Jia \emph{et al.}~\cite{vpt} proposed to add a few additional tokens, namely prompt tokens, into the input space as tunable parameters. The prompt tokens are fed into multi-head attention (MHA) together with the original tokens.
In particular, they only fine-tuned the prompt tokens while keeping the transformer backbone parameters frozen. Surprisingly, fine-tuning the prompt tokens achieved comparable or even better performance than full fine-tuning. Houlsby \emph{et al.}~\cite{adapter} inserted an adapter architecture into the Feed-Forward Network (FFN) and fine-tuned the adapter layers, aiming to adapt the pre-trained backbone weight to downstream tasks. The adapter is typically a bottleneck-like architecture consisting of a down-sample layer, a non-linear layer, and an up-sample layer. Hu \emph{et al.}~\cite{Lora} proposed a low-rank adaption approach by decomposing the increments of query transformation and value transformation into a low-rank manner and achieving higher accuracy and lower memory consumption.
Zhang \emph{et al.}~\cite{NOAH} focused on combining existing PETL methods without manual design. They trained a large supernet at first and then performed a neural architecture search on hidden dimension $h$ of Adapter, rank $r$ of LoRA, and prompt length $l$ of VPT to find the best subnet for each task using a one-shot neural architecture search algorithm \cite{2021AutoFormer}. Lian \emph{et al.}~\cite{SSF} proposed a new baseline for efficient model tuning. Taking inspiration from various normalization methods, they scaled and shifted the deep features extracted by a pre-trained model with scale and shift factors. Jie \emph{et al.}~\cite{FacT} proposed a tensorization-decomposition framework to store the weight increments, in which the weights of each ViT were tensorized into a single 3D tensor, and their increments were then decomposed into lightweight factors. In the fine-tuning process, only the factors need to be updated. Typically, the above methods insert small learnable modules into large-scale pre-trained transformer models and fine-tune these modules with downstream tasks while freezing the pre-trained transformer parameters. These methods are instructive for using pre-trained transformer backbones on various vision tasks. 


\textbf{PETuning for Few-Shot Learning.} In several practical applications, high-quality labeled data is often scarce due to expensive annotation costs and potential privacy concerns \cite{hu2021sqn, hu2022sensaturban}. Pre-trained transformer models have been successfully adapted to mitigate this limitation through techniques such as VPT \cite{vpt}, Adapter \cite{adapter}, and NOAH \cite{NOAH}, which fine-tune only a small proportion of the total parameters while maintaining competitive performance on downstream tasks. However, it remains an open question whether these techniques can be effectively applied to few-shot learning tasks, where the available training examples are even more limited. Recent studies in natural language processing have begun to explore this challenge \cite{ppt, 2021Cutting, VirtualPPT, Meta-Adapters, DBLP:conf/acl/0016TYXS022, DBLP:conf/acl/CuiHDHL22}. Building on this work, we extend the investigation to few-shot parameter-efficient tuning (PETuning) in the computer vision domain. Note that, LORA \cite{Lora} aims to maintain the identity of the output for an inserted layer when training a transformer after adding a new module, achieved by properly initializing the new module. However, we observe that LORA encounters difficulties in few-shot settings. By contrast, by incorporating our proposed PVP Tuning framework, we demonstrate a significant improvement in LORA's few-shot performance, as demonstrated in the experimental section.


\section{Proposed Method}

\subsection{Overview}

In this section, we first revisit existing PETuning techniques and then conduct exploratory experiments to verify the performance of existing PETuning techniques including VPT~\cite{vpt}, Adapter~\cite{adapter}, and LoRA~\cite{Lora} on few-shot learning tasks. Next, we propose PVP, which firstly pre-trains the tunable parameters of PETuning on a large dataset and then uses the pre-trained parameters for downstream PETuning. We summarize this section by discussing the versatility of our PVP framework.

\subsection{Revisit Parameter-Efficient Tuning Methods}
Here, we briefly recap the PETuning techniques. The key idea of PETuning is to inject a few parameters into the transformer backbone. The transformer backbone parameters are frozen to yield generalized representations learned from large-scale data and the newly inserted parameters are tunable to adapt the output distribution to specific downstream tasks. We use $\boldsymbol{F}$ to denote the vision transformer model with parameters $\theta$. For transformer architecture,

\begin{equation}
\begin{aligned}
    y = \boldsymbol{F}_\theta(x),
\end{aligned}
\label{equ_01}
\end{equation}

\noindent and the gradient is calculated as

\begin{equation}
\begin{aligned}
    g_\theta = \frac{\partial \boldsymbol{F}(\mathcal{D}; \theta) }{\partial \theta},
\end{aligned}
\label{equ_02}
\end{equation}

\noindent where $\mathcal{D}$ is large-scale training dataset. For PETuning methods, a few new parameters $\theta'$ are inserted into $\boldsymbol{F}$,

\begin{equation}
\begin{aligned}
    y = \boldsymbol{F}_{\theta,\theta'}(x),
\end{aligned}
\label{equ_03}
\end{equation}

\noindent where $\theta'$ is usually much less than $\theta$ and $\theta$ is fixed during fine tuning with only $\theta'$ learnable. The gradient update for PETuning methods is formulated as 

\begin{equation}
\begin{aligned}
    g_\theta' = \frac{\partial \boldsymbol{F}(\mathcal{D'}; \theta, \theta') }{\partial \theta'},
\end{aligned}
\label{equ_04}
\end{equation}

\noindent where $\mathcal{D'}$ is a downstream dataset for a specific task and is usually much smaller than $\mathcal{D}$.

\subsection{Exploring Few-shot Parameter-Efficient Tuning}\label{section:sec3.2}
To study the few-shot PETuning, we take VPT~\cite{vpt}, Adapter \cite{adapter}, and LoRA \cite{Lora} as examples which are tuned with limited training examples. Specifically, we tuned the VPT, Adapter, and LoRA framework with different proportions of training examples on the CUB-200-2011 dataset, reduced from 16 training samples to 8 or even 1 sample per class, validating the performance of these methods under few-shot learning settings. As shown in Figure  \ref{fig:prior_experiment}, it is clear that the performance of both three methods drops significantly when tuned with less than 4 training samples per class, and almost fails when there is only 1 training sample per class. This indicates that existing parameter-efficient tuning techniques may not be able to perform well under the few-shot learning setting.

\textbf{Analysis.} We attribute this phenomenon to the inappropriate initialization of newly added parameters because PETuning methods are used to randomly initialize these newly added parameters with a mean value of zero, meaning the newly added parameters need to learn from scratch on downstream tasks. This leads to the newly added module requiring a significant amount of data for gradient updating, which can prove challenging for downstream tasks and limits its application for few-shot tasks. To further utilize PETuning on limited data, the newly added parameters also need pre-training. Therefore, we pre-train the newly added modules to get better initialization for parameter-efficient tuning on a specific task and propose our pre-trained visual parameter-efficient tuning framework. This is intuitive as the newly added module pre-training stage can provide a good basis and the newly added parameters only require a few data to learn well on downstream few-shot tasks.

\begin{algorithm}[t]
\begin{lstlisting}[language=Python]
# For downstream prompt tuning
# type: visual prompt tuning type, "Deep" or "Shallow"
# k: visual prompt tuning tokens number

# prompt pre-training stage
# build model for prompt tokens pre-training
# type="Deep", pre-train N prompt tokens where N >= k
net=build_model(vpt_type="Deep",num_prompt_tokens=N)
for x, label in pre-train_dataloader:
    loss=net.forward(x,label)
    loss.backward()
# Prompt_tokens shape: (num_layer,num_tokens,embed_dim)
torch.save(net.Prompt_tokens,"<@\textcolor{blue}{ckpt}@>")

# pre-trained prompt tuning stage
# build model for downstream prompt tuning
net=build_model(vpt_type=type,num_prompt_tokens=k)
<@\textcolor{red}{load\_prompts}@>(net,<@\textcolor{blue}{ckpt}@>,vpt_type=type,num_token=k, load_type)
for x, label in downstream_dataloader:
    loss=net.forward(x,label)
    loss.backward()
net.test_loop()

# pre-trained prompt loading stage
def <@\textcolor{red}{load\_prompts}@>(net,<@\textcolor{blue}{ckpt}@>,vpt_type,num_token,load_type):
    if load_type=="Average":# Use averaged tokens
        checkpoint=torch.mean(<@\textcolor{blue}{ckpt}@>,dim=1)
        checkpoint=checkpoint.unsqueeze(1)
        checkpoint=checkpoint.expand(-1,num_token,-1)
    else:# Use sequential tokens
        checkpoint=<@\textcolor{blue}{ckpt}@>
    if vpt_type=="Deep":
        net.Prompt_tokens=checkpoint[:,:num_token,:]
    else:# vpt_type=="Shallow"
        net.Prompt_tokens=checkpoint[:1,:num_token,:]
\end{lstlisting}
\caption{PVP framework based on VPT, PyTorch-like.}
\label{alg:vppt}
\end{algorithm}

\definecolor{color_fgvc}{rgb}{0.99,0.37,0.0}
\definecolor{color_vtab}{rgb}{0.0,0.34,0.57}
\begin{figure*}[!htbp]
    \centering
    \includegraphics[width=0.95\linewidth]{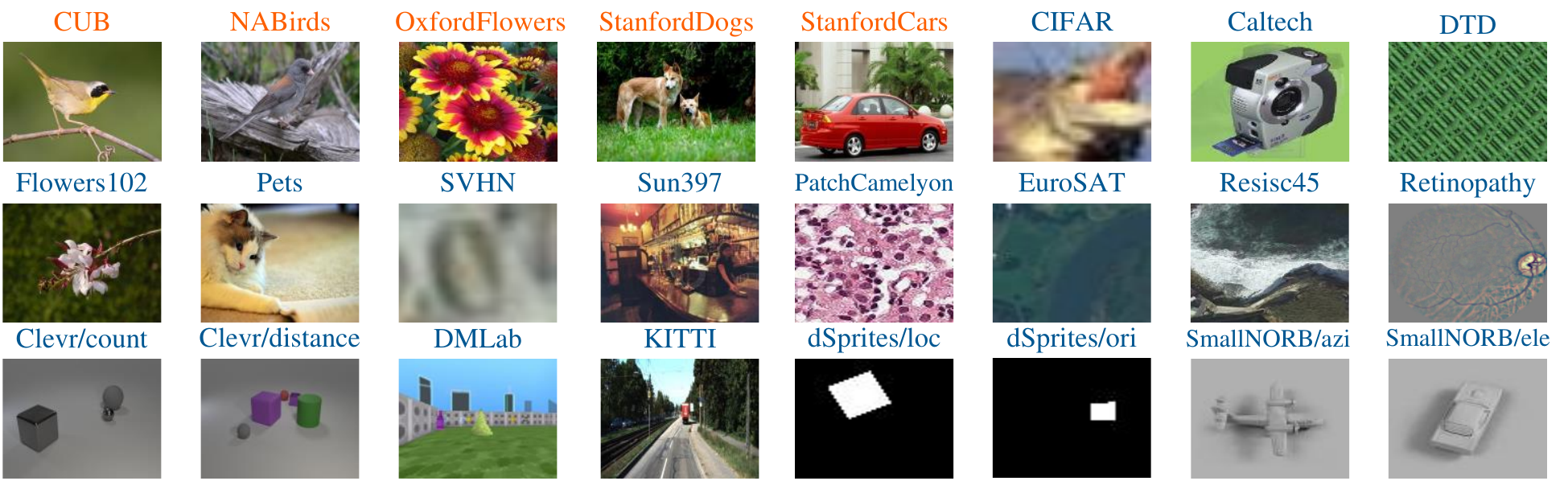}
    \caption{Examples of all classification tasks evaluated. One representative picture for each dataset in \textcolor{color_fgvc}{FGVC} and \textcolor{color_vtab}{VTAB-1k}.}
    \label{fig:dataexps}
\end{figure*}

\setlength{\tabcolsep}{4pt}
\begin{table*}[h]
\small
\begin{center}
\caption{Detailed infrmation of FGVC and VTAB-1k datasets.}
\label{table:supp_datasets}
\resizebox{0.9\textwidth}{!}{%
\begin{tabular}{l l  l l l l l}
\toprule
\textbf{Dataset}   &\textbf{Description}  & \textbf{\# Classes}    &\textbf{Train}  &\textbf{Val}  &\textbf{Test} \\ 
\midrule
\multicolumn{3}{l}{Fine-grained visual recognition tasks (FGVC)} 
\\
\cmidrule{2-6}
\quad CUB~\cite{cub}
& Fine-grained bird species recognition
&200  &\multirow{5}{*}{1/2/4/8/16 per class}
&600 &5,794
\\

\quad NABirds~\cite{nabirds}
& Fine-grained bird species recognition
&555
&&2,393	&24,633
\\

\quad Oxford Flowers~\cite{flowers}
& Fine-grained flower species recognition
&102
&&1,020	&6,149 
\\

\quad Stanford Dogs~\cite{dogs}
 &Fine-grained dog species recognition  &120 
&&1,200	&8,580 
\\

\quad Stanford Cars~\cite{cars}
& Fine-grained car classification  &196  
&&815	&8,041 
\\

\midrule

\multicolumn{3}{l}{Visual Task Adaptation Benchmark (VTAB)~\cite{VTAB}} 
\\
\cmidrule{2-6}
\quad CIFAR-100~\cite{vtab_cifar100} &\multirow{7}{*}{Natural}
&100 &\multirow{7}{*}{800/1000} &\multirow{7}{*}{200} &10,000
\\
  \quad Caltech101~\cite{vtab_caltech101} & &102 && &6,084
  \\
  \quad DTD~\cite{vtab_dtd} & &47 && &1,880
  \\
  \quad Flowers102~\cite{vtab_flowers102} & &102 && &6,149
  \\
  \quad Pets~\cite{vtab_pets} & &37 && &3,669
  \\
  \quad SVHN~\cite{vtab_svhn} & &10 && &26,032
  \\
  \quad Sun397~\cite{vtab_sun397} & &397 && &21,750
  \\
\cmidrule{2-6}

  \quad Patch Camelyon~\cite{vtab_patch_camelyon} &\multirow{4}{*}{Specialized} &2
  &\multirow{4}{*}{800/1000} &\multirow{4}{*}{200} &32,768
  \\
  \quad EuroSAT~\cite{vtab_eurosat} & &10 && &5,400
  \\
  \quad Resisc45~\cite{vtab_resisc45} & &45 && &6,300
  \\
  \quad Retinopathy~\cite{vtab_retinopathy} & &5 && &42,670
  \\

\cmidrule{2-6}
  \quad Clevr/count~\cite{vtab_clevr} &\multirow{8}{*}{Structured}
  &8
  &\multirow{8}{*}{800/1000} &\multirow{8}{*}{200} &15,000
  \\
  \quad Clevr/distance~\cite{vtab_clevr} & &6 && &15,000
  \\
  \quad DMLab~\cite{vtab_dmlab} & &6 && &22,735
  \\
  \quad KITTI/distance~\cite{vtab_kitti_dist} & &4 && &711
  \\
  \quad dSprites/location~\cite{vtab_dsprites} & &16 && &73,728
  \\
  \quad dSprites/orientation~\cite{vtab_dsprites} & &16 && &73,728
  \\
  \quad SmallNORB/azimuth~\cite{vtab_smallnorb} & &18 && &12,150
  \\
  \quad SmallNORB/elevation~\cite{vtab_smallnorb} & &9 && &12,150
\\

\bottomrule\end{tabular} }
\end{center}
\end{table*}
\setlength{\tabcolsep}{1.4pt}

\subsection{Pre-trained Visual Parameter-Efficient Tuning}

There are two stages for our Pre-trained Visual Parameter-efficient (PVP) Tuning method. As Fig.~\ref{fig:ppetuning} shows, we conduct parameter-efficient tuning on pre-train data in stage 1 and use the learned parameters to initialize the parameter-efficient tuning module for downstream tasks in stage 2.

(1) Parameter efficient tuning module pre-train stage. From Equations~\ref{equ_01}-~\ref{equ_04}, the goal of various Parameter Efficient Tuning methods is to optimize the parameters $\theta'$ using dataset $\mathcal{D'}$. However, it is difficult to directly optimize the parameters $\theta'$ when the downstream dataset $\mathcal{D'}$ is limited. Here we use another parameter efficient tuning module pre-train dataset $\mathcal{D''}$ which is larger than $\mathcal{D'}$ to pre-train the newly added parameters $\theta'$, as formulated below,

\begin{equation}
\begin{aligned}
    g_\theta' = \frac{\partial \boldsymbol{F}(\mathcal{D''}; \theta, \theta') }{\partial \theta'},
\end{aligned}
\label{equ_05}
\end{equation}

(2) Downstream parameter efficient tuning stage. We use the optimized parameters $\theta'$ in Equ.~\ref{equ_05} to initialize these in Equ.~\ref{equ_04} for our Pre-trained Visual Parameter-efficient tuning.

\definecolor{forestgreen}{rgb}{0.0, 0.27, 0.13}
\begin{table*}[h]
    \tiny
    \centering
    \caption{Quantitative results on FGVC few-shot learning.}
    \label{tab:detailed_accuracy}
    \resizebox{1\textwidth}{!}{%
    \scalebox{1.0}{
        \setlength{\tabcolsep}{1.5mm}{
        \begin{tabular}{@{}cc|cccccc@{}}
        \toprule
        \multicolumn{2}{c|}{\multirow{2}{*}{Accuracy (\%)}} & \multirow{2}{*}{CUB-200-2011} & \multirow{2}{*}{NABirds} & \multirow{2}{*}{Oxford Flowers} & \multirow{2}{*}{Stanford Dogs} & \multirow{2}{*}{Stanford Cars} & \multirow{2}{*}{Average}\\
        \multicolumn{2}{c|}{} & & & & & & \\ \midrule
        \multirow{5}{*}{FULL}
        & 16 shot & 85.12 & 79.43 & 99.20 & 72.10 & \textbf{76.91} & \textbf{82.55} \\
        & 8 shot  & 77.36 & 66.60 & 96.42 & 41.85 & \textbf{41.20} & 64.69 \\
        & 4 shot  & 60.61 & 39.10 & 94.23 & 19.80 & 23.57 & 47.46 \\
        & 2 shot  & 14.53 & 9.93 & 56.43 & 3.90 & 5.85 & 18.13 \\
        & 1 shot  & 9.44  & 2.50 & 38.61 & 1.75 & 4.17 & 11.29 \\ \midrule
        \multirow{5}{*}{VPT}          
        & 16 shot & 84.66 & 76.71 & 99.38 & 80.82 & 57.33 & 79.78 \\
        & 8 shot  & 79.10 & 64.73 & 98.75 & 77.11 & 36.31 & 71.20 \\
        & 4 shot  & 70.61 & 40.43 & 96.85 & 68.22 & 20.62 & 59.35 \\
        & 2 shot  & 53.26 & 27.94 & 92.73 & 49.02 & 8.64  & 46.32 \\
        & 1 shot  & 32.88 & 14.84 & 66.01 & 36.67 & 5.20  & 31.12 \\ \midrule
        \multirow{5}{*}{\textbf{PVP (ours)}}
        & 16 shot & \textbf{86.28} (\textcolor{forestgreen}{$\uparrow$1.62}) & \textbf{80.05} (\textcolor{forestgreen}{$\uparrow$3.34}) & \textbf{99.48} (\textcolor{forestgreen}{$\uparrow$0.10}) & \textbf{81.77} (
        \textcolor{forestgreen}{$\uparrow$0.95}) & 61.09 (\textcolor{forestgreen}{$\uparrow$3.76}) & 81.73 (\textcolor{forestgreen}{$\uparrow$1.95}) \\
        & 8 shot  & \textbf{81.53} (\textcolor{forestgreen}{$\uparrow$2.43}) & \textbf{71.78} (\textcolor{forestgreen}{$\uparrow$7.05}) & \textbf{99.02} (\textcolor{forestgreen}{$\uparrow$0.27}) & \textbf{77.81} (\textcolor{forestgreen}{$\uparrow$0.70}) & 40.57 (\textcolor{forestgreen}{$\uparrow$4.26}) & \textbf{74.14} (\textcolor{forestgreen}{$\uparrow$2.94}) \\
        & 4 shot  & \textbf{74.37} (\textcolor{forestgreen}{$\uparrow$3.76}) & \textbf{58.16} (\textcolor{forestgreen}{$\uparrow$17.73}) & \textbf{98.49} (\textcolor{forestgreen}{$\uparrow$1.64}) & \textbf{71.43} (\textcolor{forestgreen}{$\uparrow$3.21}) & \textbf{26.08} (\textcolor{forestgreen}{$\uparrow$5.46}) & \textbf{65.71} (\textcolor{forestgreen}{$\uparrow$6.36}) \\
        & 2 shot  & \textbf{62.20} (\textcolor{forestgreen}{$\uparrow$8.94}) & \textbf{53.82} (\textcolor{forestgreen}{$\uparrow$25.88}) & \textbf{96.11} (\textcolor{forestgreen}{$\uparrow$3.38}) & \textbf{62.32} (\textcolor{forestgreen}{$\uparrow$13.30}) & \textbf{14.73} (\textcolor{forestgreen}{$\uparrow$6.09}) & \textbf{57.84} (\textcolor{forestgreen}{$\uparrow$11.52}) \\
        & 1 shot  & \textbf{49.24} (\textcolor{forestgreen}{$\uparrow$16.36}) & \textbf{39.74} (\textcolor{forestgreen}{$\uparrow$24.90}) & \textbf{88.84} (\textcolor{forestgreen}{$\uparrow$22.83}) & \textbf{47.45} (\textcolor{forestgreen}{$\uparrow$10.78}) & \textbf{10.73} (\textcolor{forestgreen}{$\uparrow$5.53}) & \textbf{47.20} (\textcolor{forestgreen}{$\uparrow$16.08}) \\ \bottomrule
    \end{tabular}
    }}}
\end{table*}

\begin{table*}[h]
\scriptsize
\caption{Quantitative results on VTAB-1k transfer learning.}
\label{table:supp_vtab}
\resizebox{\textwidth}{!}{
\begin{tabular}{
c r 
rrrrrrr r!{\vrule}
rrrr r!{\vrule}
rrrrrrrr r!{\vrule}
r
}
\toprule
  &&\rotatebox{90}{\bf{CIFAR-100}}
  &\rotatebox{90}{\bf{Caltech101} }
  &\rotatebox{90}{\bf{DTD} }
  &\rotatebox{90}{\bf{Flowers102} }
  &\rotatebox{90}{\bf{Pets} }
  &\rotatebox{90}{\bf{SVHN} }
  &\rotatebox{90}{\bf{Sun397} }
  &\rotatebox{90}{\bf{Mean}}
  &\rotatebox{90}{\bf{Patch Camelyon} }
  &\rotatebox{90}{\bf{EuroSAT} }
  &\rotatebox{90}{\bf{Resisc45} }
  &\rotatebox{90}{\bf{Retinopathy} }
  &\rotatebox{90}{\bf{Mean}}
  &\rotatebox{90}{\bf{Clevr/count} }
  &\rotatebox{90}{\bf{Clevr/distance} }
  &\rotatebox{90}{\bf{DMLab}}
  &\rotatebox{90}{\bf{KITTI/distance} }
  &\rotatebox{90}{\bf{dSprites/location} }
  &\rotatebox{90}{\bf{dSprites/orientation} }
  &\rotatebox{90}{\bf{SmallNORB/azimuth} }
  &\rotatebox{90}{\bf{SmallNORB/elevation} }
  &\rotatebox{90}{\bf{Mean}}
  &\rotatebox{90}{\bf{Overall Mean}}
  \\
\midrule
\multirow{3}{*}
& \textbf{Traditional Methods}
\\
& Full Tune~\cite{vpt} &68.9 &87.7 &64.3 &97.2 &86.9 &87.4 &38.8 &75.88 &79.7 &95.7 &84.2 &73.9 &83.36 &56.3 &58.6 &41.7 &65.5 &57.5 &46.7 &25.7 &29.1 &47.64 &68.96
\\
&Linear Probe~\cite{vpt} &63.4 &85.0 &63.2 &97.0 &86.3 &36.6 &51.0 &68.93 &78.5 &87.5 &68.6 &74.0 &77.16 &34.3 &30.6 &33.2 &55.4 &12.5 &20.0 &9.6 &19.2 &26.84 &57.64
\\

\midrule
\multirow{7}{*}
&\textbf{PETuning Methods}
\\
&VPT-Shallow(ECCV'22)~\cite{vpt} &77.7 &86.9 &62.6 &97.5 &87.3 &74.5 &51.2 &76.81 &78.2 &92.0 &75.6 &72.9 &79.66 &55.5 &58.6 &40.5 &67.1 &68.7 &36.1 &20.2 &34.1 &46.98 &67.82
\\
&VPT-Deep(ECCV'22)~\cite{vpt} &\textbf{78.8} &90.8 &65.8 &98.0 &88.3 &78.1 &49.6 &78.48 &81.8 &96.1 &83.4 &68.4 &82.43 &68.5 &60.0 &46.5 &72.8 &73.6 &47.9 &32.9 &37.8 &54.98 &71.96
\\
&NOAH(arXiv'22)~\cite{NOAH} &70.7 &91.6 &68.2 &98.9 &90.2 &88.4 &54.0 &80.29 &85.9 &95.3 &84.2 &73.6 &84.75 &81.7 &63.1 &49.0 &78.5 &82.3 &45.0 &31.8 &43.5 &59.36 &74.80
\\
&SSF(Neurips'22)~\cite{SSF} &69.0 &92.6 &71.5 &99.4 &91.8 &\textbf{90.2} &52.9 &81.57 &87.4 &95.9 &87.4 &75.5 &86.55 &75.9 &62.3 &53.3 &80.6 &77.3 &54.9 &29.5 &37.9 &58.96 &75.69
\\
&FacT(AAAI'23)~\cite{FacT} &70.6 &90.6 &70.8 &99.1 &90.7 &88.6 &54.1 &80.64 &84.8 &\textbf{96.2} &84.5 &75.7 &85.30 &\textbf{82.6} &\textbf{68.2} &49.8 &80.7 &80.8 &47.4 &33.2 &43.0 &60.71 &75.55
\\
&\textbf{PVP (ours)} &76.3 &\textbf{94.4} &\textbf{73.1} &\textbf{99.7} &\textbf{92.3} &87.3 &\textbf{58.6} &\textbf{83.09} &\textbf{87.5} &95.6 &\textbf{87.4} &\textbf{76.9} &\textbf{86.84} &76.4 &64.6 &\textbf{54.6} &\textbf{82.0} &\textbf{88.0} &\textbf{58.5} &\textbf{36.2} &\textbf{52.8} &\textbf{64.13} &\textbf{78.02}
\\
\bottomrule
\end{tabular}
}
\end{table*}

\subsection{Versatility of PVP Tuning}

The key to the PVP framework is to use pre-trained prompts for downstream few-shot tasks. Given that current PETuning methods~\cite{vpt,adapter,Lora} mainly insert diverse prompt modules into vision transformers and tune these newly added modules while keeping the transformer backbone frozen. Hence, PVP is baseline-independent and can therefore apply to various PETuning methods. In this section, we study the versatility of the proposed PVP framework on VPT \cite{vpt}, Adapter \cite{adapter}, and LoRA \cite{Lora}. 

\myPara{PVP Tuning based on VPT}
The key to our approach is to use pre-trained visual prompts for prompt tuning. Specifically, we first add prompt tokens into ViT and follow VPT to initialize the prompt tokens. Next, we execute prompt tuning on ImageNet-1k with transformer backbone pre-trained on ImageNet-21k to get pre-trained prompt tokens. Finally, we load the pre-trained prompt tokens rather than tuning from scratch, before prompt tuning on downstream few-shot tasks. Algorithm~\ref{alg:vppt} shows the overall procedure of the PVP framework, including the prompt pre-training stage, pre-trained prompt loading stage and pre-trained prompt tuning stage. Notably, the number of prompt tokens in VPT varies from 1 to 200 and we directly add 200 prompt tokens into each ViT layer during the prompt pre-training stage, therefore we can load any number of pre-trained prompt tokens out of the 200 prompt tokens on downstream few-shot tasks rather than perform prompt pre-training repetitively for different prompt tokens number setting. In particular, there are two manners to load the pre-trained prompt tokens, which are listed below:

\textbf{Sequential Loading}. As the name implies, we load the pre-trained prompt tokens sequentially. For example, if there are $N$ pre-trained prompt tokens in total and we need to load $K$ pre-trained prompt tokens. In this case, we directly load the first $K$ out of the total $N$ pre-trained prompt tokens.

\textbf{Average Loading}. Different from the sequential loading manner, we use average pre-trained prompt tokens to initialize the prompt tokens. For example, if there are $N$ pre-trained prompt tokens in total and we need to load $K$ pre-trained prompt tokens, we average $N$ pre-trained prompt tokens and then expand it to $K$ tokens for loading.

\myPara{PVP Tuning based on Adapter}
Adapter insert adapter architecture to each transformer block, $$\boldsymbol{X}'\leftarrow\boldsymbol{X}+\phi(\boldsymbol{X}\boldsymbol{W}_{down})\boldsymbol{W}_{up},$$ where $\boldsymbol{X}\in \mathbb{R}^{N\times d}$ is the output of Feed-Forward Network (FFN) blocks in each transformer layer, $\phi$ is a nonlinear function, $\boldsymbol{W}_{down}\in \mathbb{R}^{d\times h}$, $\boldsymbol{W}_{up}\in \mathbb{R}^{h\times d}$ and $h << d$. We directly conduct adapter tuning on ImageNet-1k to get pre-trained $\boldsymbol{W}_{down}$ and $\boldsymbol{W}_{up}$ for each transformer block and then load these pre-trained parameters of each $\boldsymbol{W}_{down}$ and $\boldsymbol{W}_{up}$ for downstream adapter tuning.

\myPara{PVP Tuning based on LoRA}
LoRA decomposes the increments of query transformation $\boldsymbol{W}_{q}$ and value transformation $\boldsymbol{W}_{v}$ into low-rank $\boldsymbol{A}_{q/v}\in \mathbb{R}^{d\times r}$ and $\boldsymbol{B}_{q/v}\in \mathbb{R}^{r\times d}$ where $r << d$. The query and value are then computed as $$\boldsymbol{Q/V}\leftarrow\boldsymbol{XW}_{q/v}+s\cdot\boldsymbol{XA}_{q/v}\boldsymbol{B}_{q/v},$$ where $s$ is a hyper-parameter. Similar to Adapter, we first pre-train these $\boldsymbol{A}_{q/v}$ and $\boldsymbol{B}_{q/v}$ on ImageNet-1k and then use pre-trained $\boldsymbol{A}_{q/v}$ and $\boldsymbol{B}_{q/v}$ for downstream LoRA tuning.

We show experimental results about the versatility of our proposed method in Sec. \ref{section:sec4.3.2}.

\section{Experiment}

\begin{figure*}[t]
    \centering
    \includegraphics[width=0.95\linewidth]{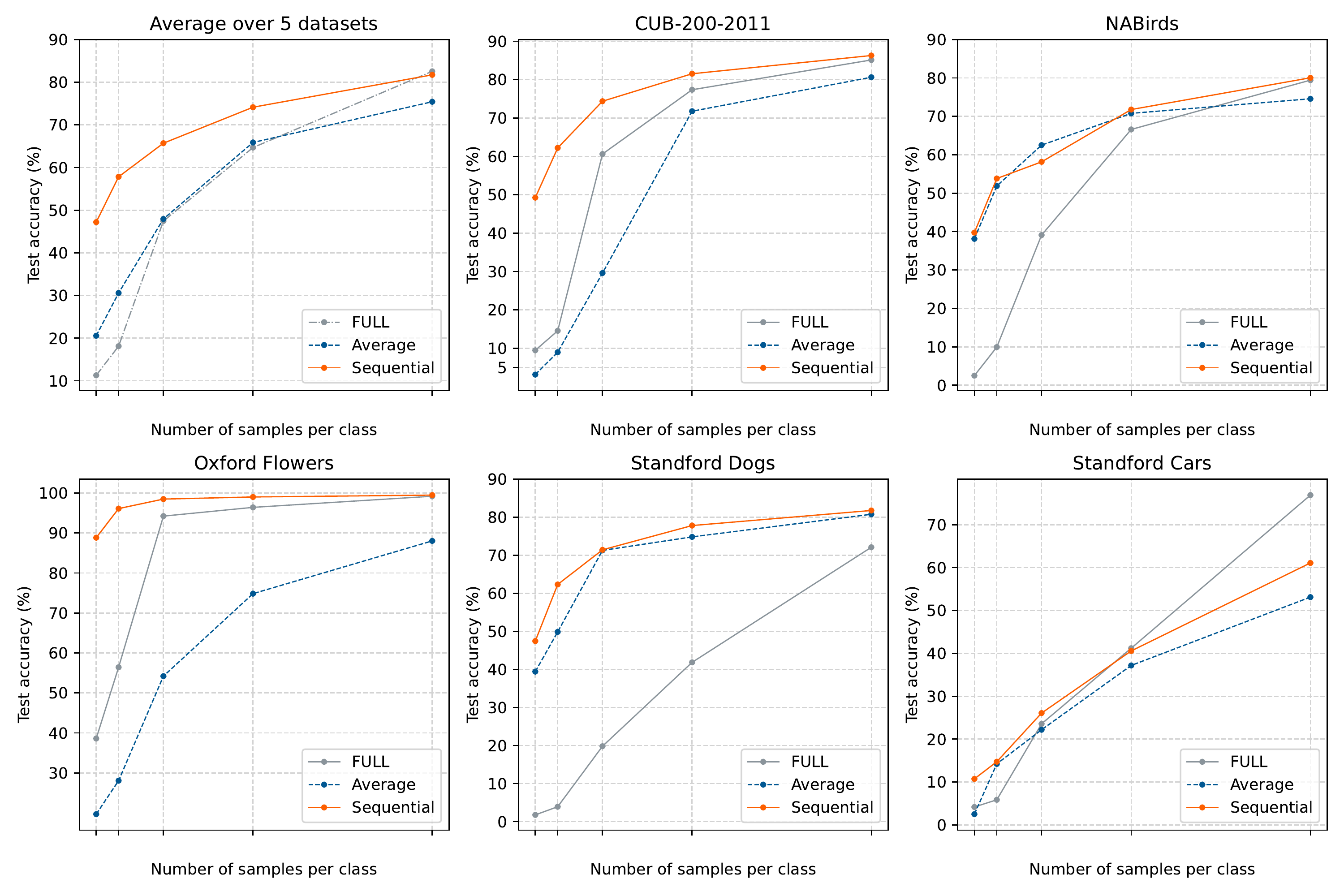}
    \caption{Result of two prompt tokens load manners in PVP(VPT). Average and sequential represent average loading and sequential loading manner.}
    \label{fig:FULL_VPPTavg_VPPT}
\end{figure*}
\begin{figure}[h]
    \centering
    \includegraphics[width=1\linewidth]{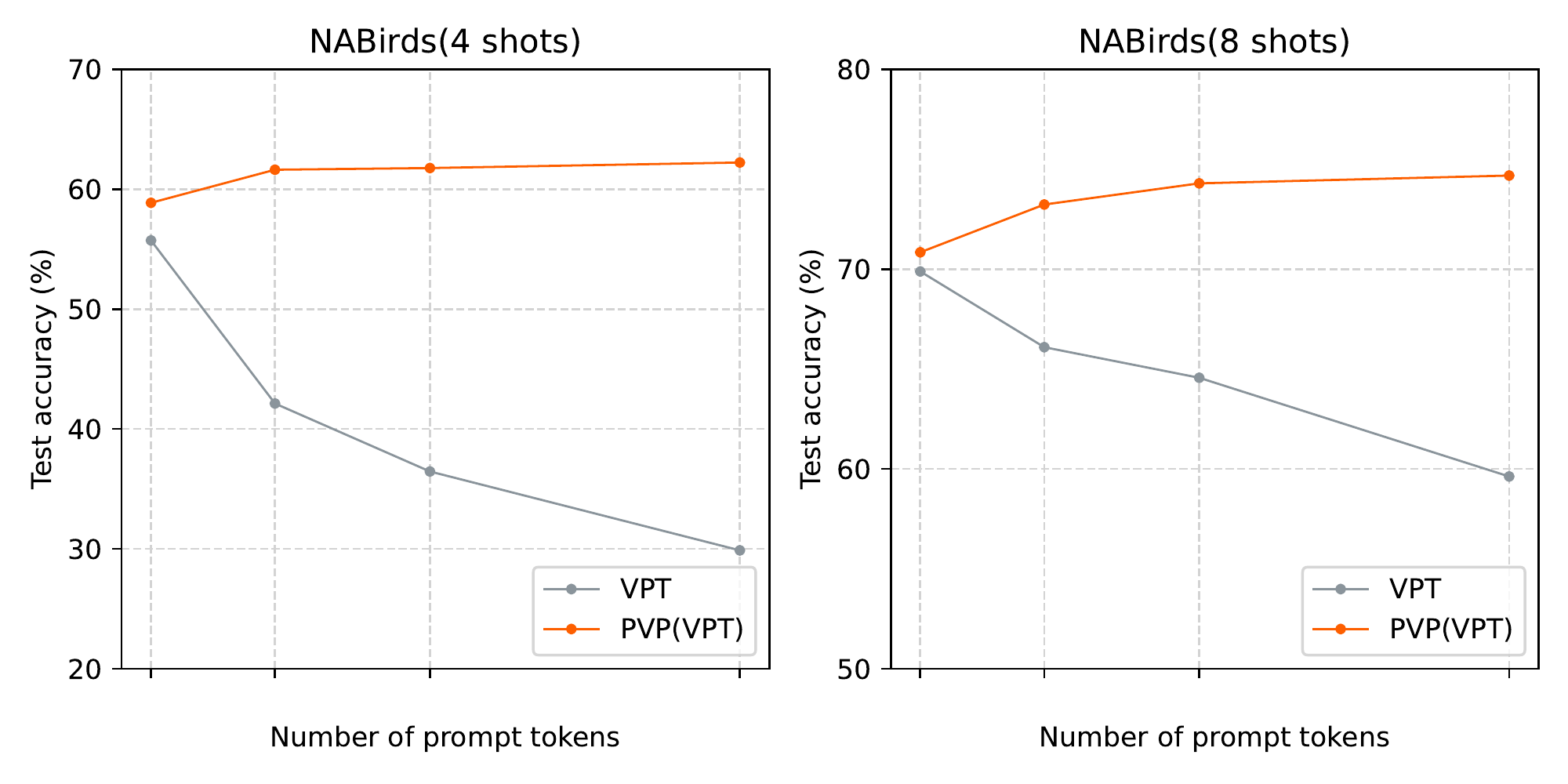}
    \caption{Test accuracy of VPT and our PVP based on VPT with different numbers of prompt tokens on NABirds dataset under 4 shots and 8 shots settings.}
    \label{fig:sensitivity}
\end{figure}

\subsection{Datasets}
For our proposed method, we evaluate the few-shot learning performance on the Fine-Grained Visual Recognition (FGVC) datasets and the transfer learning performance on the Visual Task Adaption Benchmark (VTAB-1k).

(1) FGVC contains commonly-used fine-grained visual classification datasets, which are usually used for few-shot learning, including CUB-200-2011~\cite{cub}, NABirds~\cite{nabirds}, Oxford Flowers~\cite{flowers}, Stanford Dogs~\cite{dogs} and Stanford Cars~\cite{cars}. We follow~\cite{NOAH,SSF} to use X (X=1,2,4,8,16) samples per class for few-shot image classification on these datasets.

(2) VTAB-1k~\cite{VTAB}, consisting of 19 visual classification datasets, cover data in 3 fields, including natural tasks, specialized tasks, and structured tasks. The natural task includes images in daily life. The specialized task includes images captured by specialized equipment, such as medical and satellite imagery. The structured task includes images that require semantic understanding, such as object counting. Each of the 19 datasets contains 1000 images, which reflects "1k" of the name "VTAB-1k". These datasets cover a wide range of the possible domains where downstream tasks come from, and thus the effectiveness of PETuning methods can be measured comprehensively.

Table \ref{table:supp_datasets} shows detailed information about these datasets. Examples of FGVC and VTAB-1k benchmarks are shown in Figure ~\ref{fig:dataexps}. Note that Clevr/count and Clevr/distance, dSprites/location and dSprites/orientation as well as SmallNORB/azimuth and SmallNORB/elevation are actually the same dataset but for different tasks respectively.

\subsection{Augmentation and Hyper-Parameters}
We adopt a standard image augmentation strategy during training: normalize with ImageNet means and standard deviation, randomly resize crop to 224$\times$224 and random horizontal flip for five FGVC datasets and resize to 224$\times$224 for the VTAB-1k suite. 
Following VPT~\cite{vpt}, we conduct a grid search to find the tuning-specific hyper-parameters, learning rate, and weight decay values for each task.

\begin{figure*}[h]
    \centering
    \includegraphics[width=0.95\linewidth]{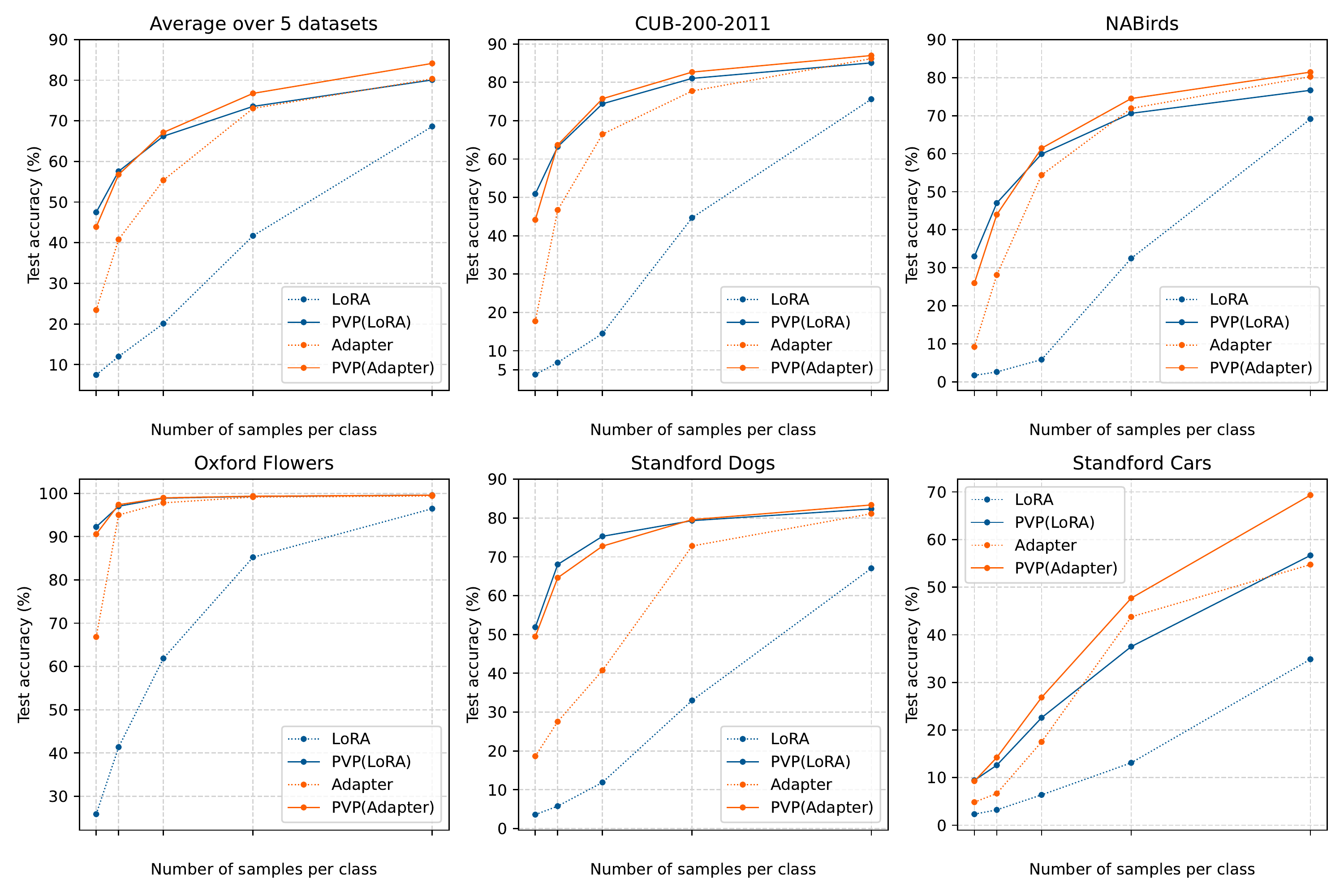}
    \caption{Result of PVP framework based on Adapter and LoRA.}
    \label{fig:other_prompt}
\end{figure*}

\subsection{Fine-Grained Few-Shot Learning}
For few-shot learning performance, we compare our methods to various competitive baselines, including VPT~\cite{vpt}, Adapter~\cite{adapter}, and LoRA~\cite{Lora}. For all baselines, we use a ViT-B/16~\cite{vit} pre-trained on supervised ImageNet-21K as the transformer backbone. All our experiments are conducted on NVIDIA A100-40GB GPUs.

\subsubsection{PVP based on VPT}
 We pre-train 200 prompt tokens on ImageNet-1k for downstream prompt tuning where the number is approximately close to the number of image
patch tokens (196) within each Multi-head Self Attention (MSA) for ViT-B/16~\cite{vit} architecture. For each downstream dataset, we follow VPT~\cite{vpt} to grid search the number of prompt tokens for a fair comparison. 

\textbf{Performance under different few-shot learning settings}. As shown in Table~\ref{tab:detailed_accuracy}, we quantitatively compare the performance achieved by FULL tuning, VPT, and the proposed PVP based on VPT under various few-shot settings on five different datasets. It can be seen that when the number of training samples is sufficient, such as 16 samples per class, PVP based on VPT reaches over 99\% test accuracy on Oxford Flowers and over 86\% test accuracy on CUB-200-2011, which is comparable to full fine-tuning using all the training samples on these two datasets. More importantly, pre-trained prompt tokens show significant performance improvement in the few-shot regime, like 1 or 2 shots per class. These results demonstrate that pre-trained prompt tokens are essential for applying large transformer models to few-shot tasks.

\textbf{Tokens load manners}. In our implementation, there are two manners to load the pre-trained prompt tokens. We study the effect of these two manners. As Figure \ref{fig:FULL_VPPTavg_VPPT} shows, loading pre-trained prompts sequentially outperforms that of averagely and we use the sequential manner as the default loading setting in the rest of this paper. We attribute the reason of low performance to the absence of positional information when loading pre-trained prompt tokens averagely, where the prompt tokens are averaged first and then expanded to the required tokens number, therefore, the positional information is missing.

\textbf{Prompt length sensitivity}. In VPT \cite{vpt}, the number of newly added prompt tokens is chosen from \{1,5,10,50,100,200\} and they use the validation set of each dataset to determine the best prompt length. We also conduct experiments on prompt tokens number to validate the sensitivity. As Figure \ref{fig:sensitivity} shows, the accuracy of VPT varies greatly (more than 25\% under 4 shots setting) when the number of prompt tokens is different, while the accuracy of our PVP framework based on VPT is more consistent and robust (less than 2\% under 4 shots setting) on NABirds dataset. We attribute the reason to the pre-training of prompt tokens since the pre-training process gives the prompt tokens better initialization and thus they can learn on limited data steadily.

\subsubsection{PVP based on Adapter and LoRA}\label{section:sec4.3.2}

Figure \ref{fig:other_prompt} shows the accuracy of the Adapter and LoRA with or w/o prompt pre-training under various shots settings. It can be seen that the proposed PVP framework brings accuracy gains on both Adapter and LoRA in few-shot learning settings, which further validates the importance of prompt pre-training as well as the versatility of the PVP framework.

\subsection{VTAB-1k Transfer Learning}
For transfer learning performance, we compare our methods to various PETuning methods, including VPT-Deep~\cite{vpt}, VPT-Shallow~\cite{vpt}, NOAH~\cite{NOAH}, SSF~\cite{SSF} and FacT~\cite{FacT}. We use a ViT-B/16~\cite{vit} pre-trained on supervised ImageNet-21K as the transformer backbone and use VPT as our baseline. Following VPT~\cite{vpt}, we search the number of prompt tokens from $\{$1,10,50,100,200$\}$ for each dataset. For convenience, we directly use prompt tokens pre-trained on ImageNet-1k for both natural tasks, specialized tasks, and structured tasks. All our experiments are conducted on NVIDIA A100-40GB GPUs.

Experimental results are shown in Table ~\ref{table:supp_vtab}, from which we can see that:

1) Our PVP(VPT) reaches comparable performance with respect to previous state-of-the-art PETuning methods. On VTAB-1k benchmark, PVP(VPT) achieves the highest accuracy on 16 datasets out of 19 datasets in total and achieves 1.52$\%$, 0.29$\%$ and 3.42$\%$ average accuracy improvement on natural tasks, specialized tasks, and structured tasks, respectively.

2) Though we use prompts pre-trained on ImageNet-1k, which is mainly about natural images, our PVP framework based on VPT also performs well on specialized tasks and structured tasks.

\section{Conclusion}
In this paper, we study recent Parameter-Efficient Tuning (PETuning) methods and first observe that current PETuning methods perform poorly in the few-shot scenario. Then, we propose Pre-trained Visual Parameter-efficient (PVP) Tuning, a conceptually simple and intuitive framework to leverage large-scale pre-trained transformer models for few-shot tasks. The key to our method is to pre-train the prompt modules of recent PETuning methods, enabling better initialization for downstream PETuning. Extensive experiments on VPT, Adapter, and LoRA show the effectiveness and versatility of the PVP framework in terms of few-shot learning. Besides the few-shot capability, PVP also shows comparable transfer learning ability to recent PETuning methods. On VTAB-1k benchmark, PVP achieves state-of-the-art results on 16 out of total 19 datasets and improves the average accuracy of 3.34$\%$, 2.85$\%$, and 3.66$\%$ for VTAB-Natural, VTAB-Specialized and VTAB-Structured, respectively. We hope our work can inspire future research on more efficient and lightweight utilization of large vision models. However, Pre-trained Visual Parameter-efficient (PVP) Tuning is an empirical method with experiments proof currently. Though state-of-the-art results were achieved on FGVC and VTAB-1k benchmarks, the theoretical interpretation behind PVP is still under exploration.

\myPara{Acknowledgement}
This work is supported by the National Natural Science Foundation of China under Grants 62006241, and 61902415.

\newpage

\bibliographystyle{IEEEtran}
\bibliography{egbib}

\vfill

\end{document}